\documentclass{article}
\usepackage{spconf,amsmath,graphicx}
\usepackage{multirow}
\usepackage{booktabs}
\usepackage{subfigure}

\title{Data Poisoning Attack Aiming the Vulnerability of Continual Learning}
\name{Gyojin Han$^{1}\sthanks{Equal contribution.}$, Jaehyun Choi$^{1\ast}$, Hyeong Gwon Hong$^{2}$, and Junmo Kim$^{1}$}
\address{$^{1}$School of Electrical Engineering, KAIST, South Korea\\
$^{2}$Kim Jaechul Graduate School of AI, KAIST, South Korea}
\begin{document}
\maketitle
\begin{abstract}
Generally, regularization-based continual learning models limit access to the previous task data to imitate the real-world constraints related to memory and privacy.
However, this introduces a problem in these models by not being able to track the performance on each task.
In essence, current continual learning methods are susceptible to attacks on previous tasks.
We demonstrate the vulnerability of regularization-based continual learning methods by presenting a simple task-specific data poisoning attack that can be used in the learning process of a new task.
Training data generated by the proposed attack causes performance degradation on a specific task targeted by the attacker.
We experiment with the attack on the two representative regularization-based continual learning methods, Elastic Weight Consolidation (EWC) and Synaptic Intelligence (SI), trained with variants of MNIST dataset.
The experiment results justify the vulnerability proposed in this paper and demonstrate the importance of developing continual learning models that are robust to adversarial attacks.
\end{abstract}


\section{Introduction}
\label{sec:intro}
Humans can continuously learn new concepts throughout their lifetime while retaining previously learned knowledge.
In contrast, neural networks trained on a new task that lies in a different distribution from previous tasks, suffer from performance degradation on previous tasks by losing information. 
This phenomenon, known as \emph{catastrophic forgetting} \cite{french1999catastrophic}, happens due to one of the limitations of the neural networks: train and test data must be in the same distribution for the network to perform well.
To ease \emph{catastrophic forgetting}, continual learning (also termed lifelong or incremental learning) \cite{chen2018lifelong} seeks to obtain a single model that works well on all of the learned tasks while incrementally training the model with access only to the current training task data.

\begin{figure}[t!]
\begin{center}
\centerline{
\includegraphics[width=0.87\columnwidth]{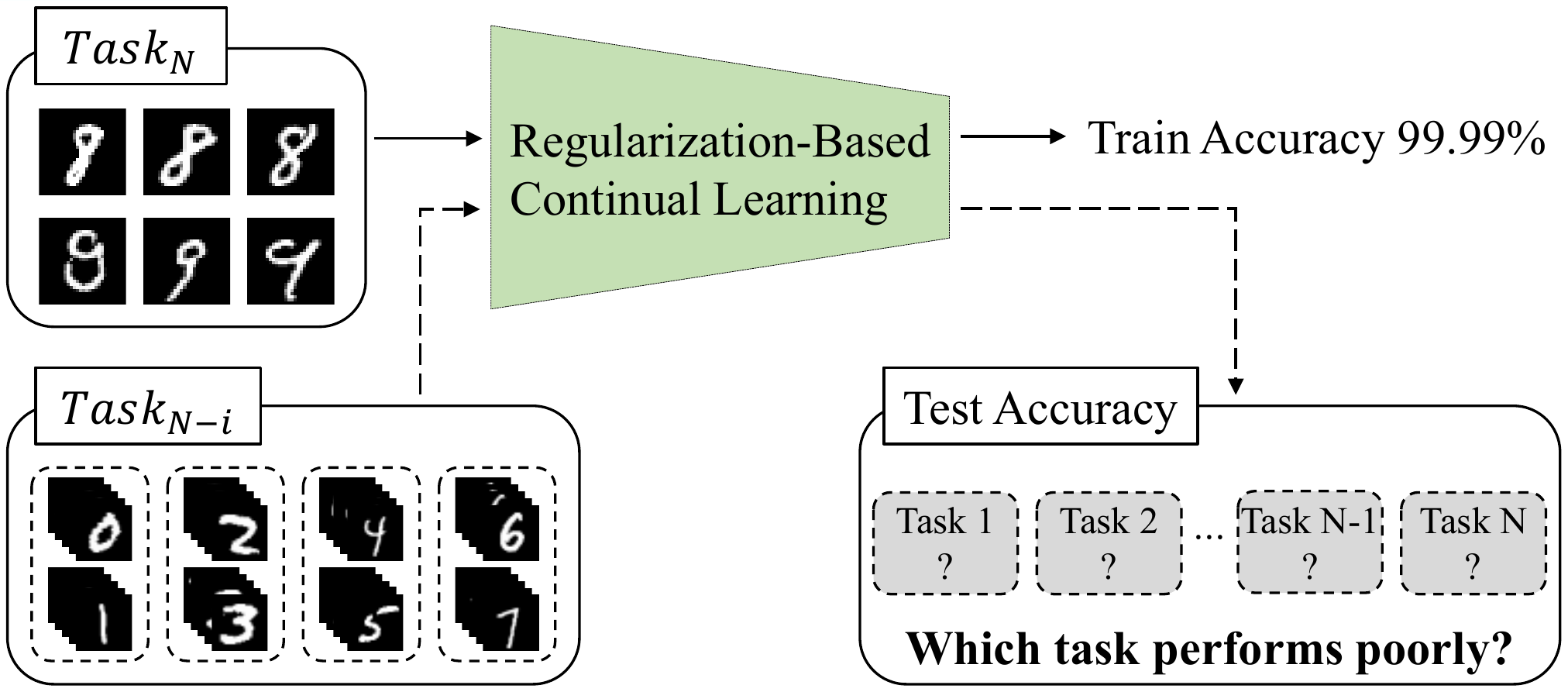}
}
\caption{The regularization-based continual learning methods do not hold the data from the previous data. The model will be validated only by the train accuracy before publishing the model to the users. However, the train accuracy does not show the model's performance on each past task, thereby being unable to detect whether attacks are done to the model or not.}
\label{explain-fig}
\end{center}
\end{figure}

Recent continual learning methods have overcome \emph{catastrophic forgetting} showing remarkable performance in both past and current tasks.
However, in a real-world situation, it is impossible to verify whether the model still works well on the past tasks since the previous data is not available due to memory and privacy problems.
In other words, train accuracy is the only option to validate the performance of the continual learning models, thereby blindly utilizing the model even when the performance on one of the tasks is low as demonstrated in the Fig.~\ref{explain-fig}.
The difficulty of tracking the reliability of the model on past tasks is a serious problem especially when the attack is done to a specific task that the model has learned in the past as it would not affect the train accuracy.
Despite such a problem, adversarial attacks \cite{43405} and defenses are not actively discussed in the field of continual learning.

In this paper, we bring forward the aforementioned problem in continual learning and experimentally justify the vulnerability with a simple task-specific data poisoning attack.
The attack is designed to not affect the training accuracy of the model to be indistinguishable when training.
More specifically, we generated adversarial data \cite{feng2019learning} during training that behaves in a new way, to work in continual learning.
The generated adversarial data of a new task only severely degrades the performance of the targeted task.

\begin{figure*}[t!]
\begin{center}
\centerline{
\includegraphics[width=1.5\columnwidth]{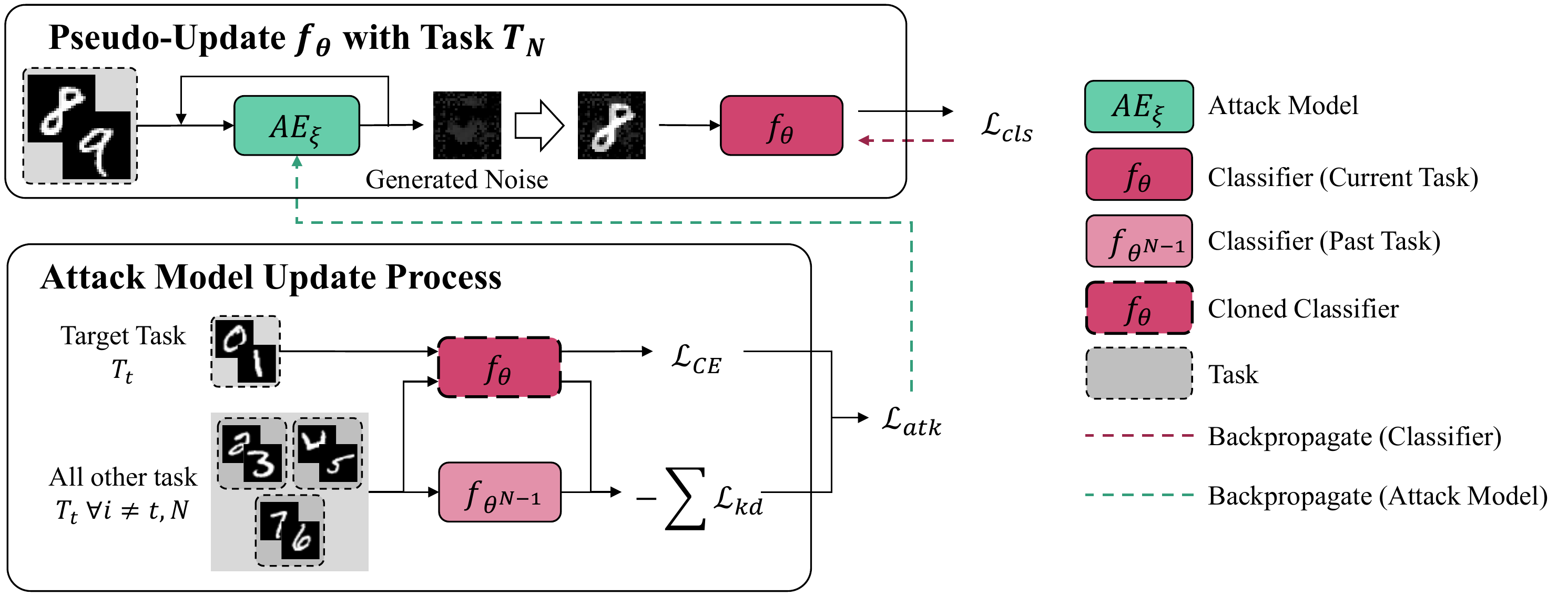}
}
\caption{The attack model loss $\mathcal{L}_{atk}$ is calculated from the pseudo-updated classifier $f_\theta$ and classifier from the past task $f_{\theta^{N-1}}$. The gradient is transmitted to the attack model $AE_\xi$ through partial differentiation. The attack model $AE_\xi$ then generates noise that gets added to the original data from the current task which updates the pseudo-updated classifier $f_\theta$ with cross-entropy loss.
}
\label{method-fig}
\end{center}
\end{figure*}

\section{Related Works}
\textbf{Continual Learning.}
Continual learning has three streams: rehearsal-based methods \cite{Rebuffi_2017_CVPR, NIPS2017_f8752278}, architecture-based methods \cite{rusu2016progressive, NEURIPS2018_cee63112, fernando2017pathnet, Mallya_2018_CVPR, pmlr-v80-serra18a}, and regularization-based methods \cite{Kirkpatrick3521, zenke2017continual}.
Among them, regularization-based methods add a regularization loss term to the loss function when learning a new task to reduce the amount of change in parameters that are important for classifying the previous tasks. In this paper, we consider the regularization-based method setting, having no access to data from past tasks during the training process for the new task.
As there are no previous task data available, the regularization-based continual learning method cannot track the performance of the past task hence relying solely on the training accuracy of the current task when deploying the model.
This opens the chance for data poisoning to easily attack the continual learning methods on previous tasks without getting detected during training time.
We demonstrate how vulnerable the continual learning methods are with a simple task-specific data poisoning attack.

\textbf{Adversarial Attack.}
First introduced in \cite{43405}, adversarial examples refer to samples with a very small perturbation, usually imperceptible by human eyes but noticeable by machine learning models, creating a gap in the inference results between them.
Although there are image-agnostic methods~\cite{moosavi2017universal, borkar2020defending}, this paper deals primarily with image-dependent adversarial attacks which are methods for generating such adversarial examples.
It can be categorized into test time adversarial attacks \cite{43405, madry2017towards, 7958570} and data poisoning attacks \cite{feng2019learning}.
Test time adversarial attacks generate images during inference and aim for incorrect inference results whereas data poisoning attacks generate images when training to make the model be trained erroneously.
In this work, we propose a data poisoning attack for continual learning that adds perturbation to the training data of a new task so that the victim continual learning model loses information about the previous task specified by the attacker while learning the new task.

\section{Proposed Method}
\label{sec:method}

In this section, we describe how an attacker generates adversarial data that cause a continual learning model to lose information on a previous task.
More specifically, the model trainer is provided with adversarial training data that does not affect the train accuracy of a new task to prevent the trainer from detecting the attack.
Moreover, as the model user might lose trust in the model if it does not work for all previous tasks, we propose an attack that only affects the performance of a specific task that the attacker intends to target.
We assume a regularization-based method setting in which training data $\{D_{1},{\ldots},D_{N}\}$ corresponding to $N$ tasks $\{T_{1},{\ldots},T_{N}\}$ is provided sequentially. Training on the data of the new task proceeds without access to previous data.
The attacker is provided with the training data $\{D_{1},{\ldots},D_{N}\}$ and a classifier $f_{\theta^{N-1}}$ trained up to $(N-1)$-th task. 
The goal of the attacker is to make the victim classifier lose knowledge about a target task $T_t$ while being trained well on the new task, where the $t$-th task is the target task. We emphasize that the attacker only slightly modifies the training data of the new task $T_N$ for flawless training on $T_N$ while losing information of $T_t$. 

\subsection{The attack process}
We use an attack model $AE_{\xi}$ with an encoder-decoder structure to manipulate the training data of a new task into adversarial data.
It takes the clean training data of the new task $D_N$ as input and generates noise that is bounded by $(-\epsilon, \epsilon)$.
The generated noise is added to $D_N$, and becomes the adversarial training data ${D_N}^{\prime}$ that can degrade the performance of the continual learning model.
To train the attack model $AE_{\xi}$, we use the optimization method proposed by \cite{feng2019learning} with modifications in training process.
The modified training process repeats the following two steps, (1) recording the trajectories of a temporary model by updating it with adversarial data, and (2) training the attack model along the trajectories by pseudo-updating the recorded parameters.
Fig. \ref{method-fig} illustrates the second step of training process of the attack model $AE_\xi$.

\textbf{Recording the trajectories of a temporary model.}
For the optimization of $AE_\xi$, we need to approximate the trajectories of $f_{\theta^{N-1}}$ when it learns the adversarial image generated by $AE_\xi$. Therefore, we use a temporary model $f_{\theta}$ for episodic training. $f_{\theta}$ is trained with adversarial training data ${D_N}^{\prime}$ generated by fixed $AE_{\xi}$. At this time, $f_{\theta}$ should be trained with a continual learning approach, as in the actual situation. Therefore, the loss for $f_\theta$, $\mathcal{L}_{cls}$ is:
\begin{equation}
\mathcal{L}_{cls}=\mathcal{L}_{CE}\left({f_{\theta}(x^{N}+AE_{\xi}(x^{N}))}, y^{N}\right) +  \Omega^{N-1}_m\\
\end{equation}
where $\Omega^{N-1}_m$ is the regularization term of the continual learning method $m$, and ($x^N$, $y^N$) is the mini-batch of the training data of the $N$-th task.
Through the loss $\mathcal{L}_{cls}$, the parameters $\theta$ are updated and recorded as follows:
\begin{equation}
\theta \leftarrow \theta-\alpha_{f}\cdot \nabla_{\theta} \mathcal{L}_{cls}
\end{equation}
where $\alpha_{f}$ is the learning rate for $f_{\theta}$.

\textbf{Training the attack model along the trajectories.} We pseudo-update recorded $f_\theta$ using the image generated by $AE_\xi$ with same loss $\mathcal{L}_{cls}$ as in step (1). Cross-entropy loss is calculated from the data of the target task $D_t$ using pseudo-updated $f_\theta$, and gradient values are transmitted to $AE_\xi$ through the loss. Then $AE_\xi$ can be updated with the gradient ascent. However, owing to the relevance between tasks, if an attack on the target task $T_t$ is attempted without any restrictions, the performance of the classifier against the other tasks involved will also be reduced. Therefore, appropriate constraints are required to maintain the classifier performance for other tasks. We want to preserve the outputs of inferences of $f_{\theta^{N-1}}$ for other tasks even if it is trained using adversarial training data. Therefore, we add the knowledge distillation loss term \cite{44873} to the loss function for training $AE_{\xi}$. The knowledge distillation loss between the outputs of $f_{\theta^{N-1}}$ and $f_\theta$, prevents adversarial data from affecting the inferences on other tasks. To calculate knowledge distillation loss, we sampled the data from the training data of each task. The knowledge distillation loss term for the $k$-th task $T_k$ is:
\begin{equation}
\mathcal{L}_{kd} = \alpha_{kd}\cdot{T^{2} \mathcal{L}_{KLD}\left(\sigma\left(\frac{f_{\theta}(x^{k})}{T}\right), \sigma\left(\frac{f_{\theta^{N-1}}(x^{k})}{T}\right)\right)}
\end{equation}
where $\alpha_{kd}$ is the balancing parameter of the knowledge distillation loss, $\mathcal{L}_{KLD}$ is the KL-divergence loss,  $T$ is the temperature parameter, and $\sigma(\cdot)$ is the softmax function. Because $AE_{\xi}$ is trained via gradient ascent, the knowledge distillation loss term for all tasks except $T_t$ and $T_N$ is subtracted from the cross-entropy loss. The loss for $AE_{\xi}$, including the knowledge distillation loss term is:
\begin{equation}
\mathcal{L}_{atk}=\mathcal{L}_{CE}\left({f_{\theta}(x^{t})}, y^{t}\right)-\sum_{i\neq{t,N}}\mathcal{L}_{kd}(f_{\theta}(x^i), f_{\theta^{N-1}}(x^i))
\end{equation}
Finally, the parameters of the attack model $\xi$ are updated as follows.
\begin{equation}
\xi \leftarrow \xi+\alpha_{AE}\cdot \nabla_{\xi} \mathcal{L}_{atk}
\end{equation}

\section{EXPERIMENTS}

To effectively validate the vulnerability of the proposed problem in continual learning, the experiment setting is chosen with the following considerations: 1) utilized continual learning methods must successfully alleviate the catastrophic forgetting, 2) the performance of all the past tasks must be high to show how easily the task-specific data poisoning drops the performance of a specific task.
Accordingly, we experiment on two continual learning methods, EWC and SI, and two variants of MNIST \cite{lecun1998mnist} dataset, permuted MNIST \cite{serra2018overcoming} and split MNIST \cite{zenke2017continual}.

\begin{figure}[t!]

\centerline{
  \subfigure[Permuted MNIST]{
    \includegraphics[width=0.05\textwidth]{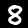}
    \includegraphics[width=0.05\textwidth]{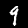}
    \includegraphics[width=0.05\textwidth]{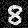}
    \includegraphics[width=0.05\textwidth]{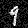}
    }
  \subfigure[Split MNIST]{
    \includegraphics[width=0.05\textwidth]{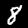}
    \includegraphics[width=0.05\textwidth]{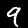}
    \includegraphics[width=0.05\textwidth]{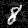}
    \includegraphics[width=0.05\textwidth]{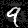}
    }
    }
  \caption{Example of clean samples (left two samples) and adversarial samples (right two samples) for (a) permuted MNIST, and (b) split MNIST. The perturbations on the samples appear differently due to the differences in the target task.}
  \label{adversarial-samples}
\end{figure}

\subsection{Experiment details.}
The attack model consists of an encoder and decoder.
The encoder has 3 $\times$ 3 convolution layers with 16, 64, and 128 channels, and the decoder has a 5 $\times$ 5 convolution layer with 128 channels and a 2 $\times$ 2 convolution layer with 64 channels.
We train the attack model with Adam optimizer for 10 epochs with learning rate of 0.0001 and batch size of 256.
The weight $\epsilon$ which determines the magnitude of the generated noise when adding to the clean sample is set to 0.2.
% All of the experiments including baseline model are trained with SGD optimizer.

\begin{figure*}[t!]
\centerline{
\includegraphics[width=2.0\columnwidth]{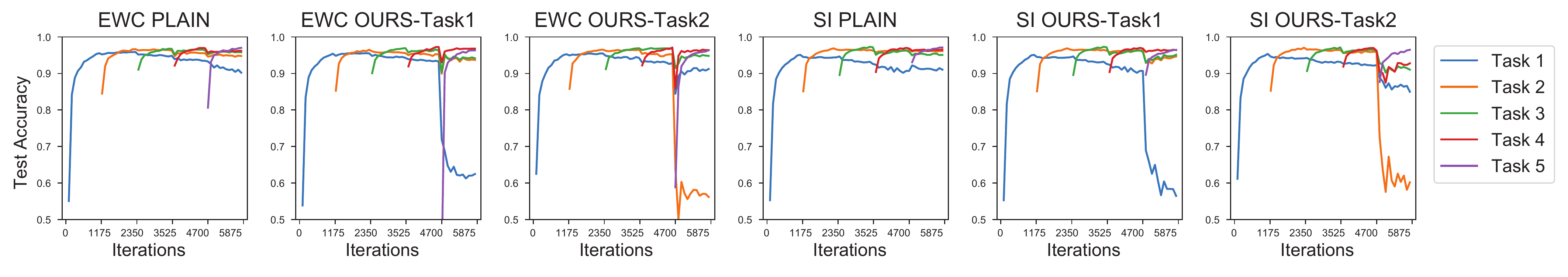}
}
\caption{Test accuracy on permuted MNIST for five tasks using EWC and SI. The graphs named 'PLAIN' show the effect of continual learning when no attacks are applied. The graphs named 'OURS-Task1' and 'OURS-Task2' show the test accuracy when $T_1$ and $T_2$ were attacked by our method, respectively. Best viewed zoomed in.}
\label{acc-graph}
\end{figure*}

\begin{table}[t!]
\centering
{\footnotesize
\resizebox{\columnwidth}{!}{
\begin{tabular}{c|cc|ccccc}
\noalign{\smallskip}\noalign{\smallskip}\toprule
Dataset & \multicolumn{2}{c|}{Method} & $T_1$ & $T_2$ & $T_3$ & $T_4$ & $T_5$\\
\midrule
\multirow{9}{*}{Permuted MNIST} &\multicolumn{2}{c|}{SGD} & 0.8059 & 0.8817 & 0.9306 & 0.9558 & 0.9520\\
\cmidrule{2-8}
% 0.9029, 0.9493, 0.957, 0.9623, 0.97
& \multirow{4}{*}{EWC} & Plain & 0.9029 & 0.9493 & 0.9570 & 0.9623 & 0.9700\\
& & Noise & 0.8657 & 0.9356 & 0.9273 & 0.9687 & 0.9617 \\
& & Ours-$T_1$ & \textbf{0.6005} & 0.9344 & 0.9360 & 0.9666 & 0.9620\\
& & Ours-$T_2$ & 0.9054 & \textbf{0.5625} & 0.9486 & 0.9639 & 0.9619\\
\cmidrule{2-8}
& \multirow{4}{*}{SI} & Plain & 0.9102 & 0.9605 & 0.9529 & 0.9658 & 0.9717\\
& & Noise & 0.9152 & 0.9561 & 0.9324 & 0.9589 & 0.9646\\
& & Ours-$T_1$ & \textbf{0.5190} & 0.9364 & 0.9386 & 0.9592 & 0.9589\\
& & Ours-$T_2$ & 0.8584 & \textbf{0.5618} & 0.9013 & 0.9183 & 0.962\\
\midrule
\multirow{9}{*}{Split MNIST} &\multicolumn{2}{c|}{SGD} & 0.4019 & 0.5901 & 0.1441 & 0.9084 & 0.9844\\
\cmidrule{2-8}
& \multirow{4}{*}{EWC} & Plain & 0.4317 & 0.7424 & 0.1254 & 0.9305 & 0.9813\\
& & Noise & 0.4132 & 0.6459 & 0.1660 & 0.8676 & 0.9803 \\
& & Ours-$T_1$ & \textbf{0.3825} & 0.5843 & 0.1596 & 0.9592 & 0.9773\\
& & Ours-$T_2$ & 0.4463 & \textbf{0.5563} & 0.2006 & 0.9350 & 0.9692\\
\cmidrule{2-8}
& \multirow{4}{*}{SI} & Plain & 0.4790 & 0.8242 & 0.3010 & 0.9728 & 0.9531\\
& & Noise & 0.4643 & 0.7919 & 0.3116 & 0.9733 & 0.9531\\
& & Ours-$T_1$ & \textbf{0.3939} & 0.8095 & 0.4242 & 0.8454 & 0.9576\\
& & Ours-$T_2$ & 0.4577 & \textbf{0.7767} & 0.4248 & 0.8348 & 0.9551\\
\bottomrule
\end{tabular}
}
}
\caption{Final accuracy of the victim classifier $f_\theta$ to evaluate the performance of our attack.}
\label{acc-table}
\end{table}

\subsection{Results}

Adversarial samples made by our task-specific data poisoning attack method can be seen in Fig.~\ref{adversarial-samples}.
Tab.~\ref{acc-table} shows the final accuracy of the continual learning model after training for all tasks is completed.
SGD in Tab.~\ref{acc-table} is the baseline method.
The `Plain' results of EWC and SI show the effectiveness of each method by being higher than baseline results.
Additionally, random uniform noise added to the clean sample is denoted by `Noise'.
`Ours-T1' and `Ours-T2' denote our task-specific data poisoning attacks done on task 1 and task 2, respectively.

As can be seen in Fig. \ref{acc-graph}, the noise created by the proposed attack caused the victim classifier to forget the knowledge about $T_1$ as it learns $T_5$.
This proves the existence of adversarial data that causes much more severe catastrophic forgetting compared with clean data as the results of `Noise', `Ours-T1', and `Ours-T2' in Tab.~\ref{acc-table} show.
For split MNIST, the results of non-targeted tasks are not stable.
This is due to the digits in MNIST dataset sharing many morphological features at the image patch level (e.g. 1\&7, 3\&8).
Therefore, attacking a task inevitably affects the parameters of the other tasks decreasing or increasing the performance of non-targeted tasks in split MNIST.
The targeted task result of `Ours-T1' and `Ours-T2' being lower than the `Noise' shows that our attack method successfully attacks the targeted task.
More importantly, the performance of non-targeted tasks stays in the reasonable range in line with the continual learning methods.
This demonstrates that the attack on the target task cannot be noticed until inference on the target task occurs even for the deployed models.
Furthermore, this shows that highly covert attacks on past tasks are possible because of the untraceable accuracy problem for past tasks in continual learning.

\subsection{Ablation study}
\textbf{Knowledge distillation loss.} We calculated the backward transfer \cite{lopez2017gradient} of the new task for the remaining tasks, except the target task to confirm the effectiveness of the knowledge distillation loss term. The backward transfer $\mathrm{B}$ was calculated as follows:
\begin{equation}
\mathrm{B}=\frac{1}{N-2} \sum_{k\neq{t,N}} R_{N, k}-R_{k, k}
\end{equation}
where $R_{i,j}$ is the test accuracy of the classifier on $T_j$ just after trained with $T_i$. 
\begin{table}[t]
\centering
{\small
\begin{tabular}{c|ccc}
\noalign{\smallskip}\noalign{\smallskip}\toprule
 & Plain & \begin{tabular}{@{}c@{}}Ours\\ w/o $\mathcal{L}_{kd}$\end{tabular}  & \begin{tabular}{@{}c@{}}Ours\\ with $\mathcal{L}_{kd}$\end{tabular} \\
\midrule
EWC & -0.0129 & -0.0717 & -0.0195\\
\midrule
SI & -0.0087 & -0.0600 & -0.0237\\
\bottomrule
\end{tabular}
}
\caption{Backward transfer of $T_5$ for $T_2$, $T_3$, and $T_4$ on permuted MNIST.}
\label{bwt-table}
\end{table}

As shown in Tab.~\ref{bwt-table}, by placing a constraint on using the knowledge distillation loss when training the attack model, the increase in negative backward transfer owing to the attack is  significantly reduced.

\section{Conclusion}
\label{sec:conclusion}
We reported weakness in continual learning caused by not having access to the data of previous tasks.
This hinders performance tracking of previous tasks, which might reduce the reliability of the continual learning models and pose a serious problem for detecting adversarial attacks.
In this regard, we propose a task-specific data poisoning attack scenario that this vulnerability could cause.
The proposed attack degrades the performance of the continual learning model on the targeted task by adding perturbations to the training data of a new task.
We highlight the importance of developing robust continual learning models by demonstrating the existence of adversarial data that causes the loss of knowledge of past tasks and suggest a simple attack scenario. 
\newline\newline\noindent\textbf{Acknowledgements.} This work was conducted by Center for Applied Research in Artificial Intelligence (CARAI) grant funded by DAPA and ADD (UD230017TD).

\bibliographystyle{IEEEbib}
\bibliography{references}

\end{document}